%% file: time2graph.tex
\def\year{2020}\relax
\pdfoutput=1 
\documentclass[letterpaper]{article} 
\usepackage{aaai20}  
\usepackage{times}  
\usepackage{helvet} 
\usepackage{courier}  
\usepackage[hyphens]{url}  
\usepackage{graphicx} 
\usepackage{graphicx}  
\usepackage{url}            
\usepackage{booktabs}       
\usepackage{amsfonts}       
\usepackage{nicefrac}       
\usepackage{microtype}      
\usepackage{comment}
\usepackage{booktabs} 
\usepackage{nicefrac}       
\usepackage{microtype}      
\usepackage{url}  
\usepackage{bm}
\usepackage{enumitem}
\usepackage{amssymb}
\usepackage{amsmath}
\usepackage{multirow}
\usepackage{subcaption}
\usepackage{changepage}
\usepackage{booktabs}
\usepackage{array}
\usepackage{diagbox}
\usepackage{rotating}
\usepackage[ruled]{algorithm2e}
\usepackage{algorithmic}
\usepackage{eqparbox}
\usepackage{makecell}
\newtheorem{definition}{Definition}

\graphicspath{{figure/}}

\newcommand{\hide}[1]{} 

\newcommand{\eqnref}[1]{Eq.~(\ref{#1})}

\newcommand{\secref}[1]{Sec.~\ref{#1}}
\newcommand{\tableref}[1]{Table~\ref{#1}} 
\newcommand{\algref}[1]{Algorithm~\ref{#1}}
\newcommand{\modelname}{\textit{Time2Graph}\xspace}
\newcommand{\sgdata}{ECR\xspace}
\newcommand{\teledata}{NTF\xspace}
\newcommand{\ucreqk}{EQS\xspace}
\newcommand{\ucrworms}{WTC\xspace}
\newcommand{\ucrstra}{STB\xspace}
\newcommand{\numbaseline}{17~}

\newcommand{\figref}[1]{Fig.~\ref{#1}} 
\newcommand{\pluseq}{\mathrel{+}=}
\newcommand{\vpara}[1]{\vspace{0.01in}\noindent\textbf{#1 }}
\newcolumntype{P}[1]{>{\centering\arraybackslash}p{#1}}

\newcommand{\citet}[1]{\citeauthor{#1} \shortcite{#1}} 
\newcommand{\citep}{\cite}

\makeatletter
\renewcommand\footnoterule{%
	\kern-3\p@
	\hrule\@width.4\columnwidth
	\kern2.6\p@}
\makeatother

\title{Time2Graph: Revisiting Time Series Modeling with \\ Dynamic Shapelets}
\author{%
	Ziqiang Cheng\textsuperscript{\dag}, Yang Yang\textsuperscript{\dag}\thanks{Corresponding author: Yang Yang, yangya@zju.edu.cn}, Wei Wang\textsuperscript{\ddag}, Wenjie Hu\textsuperscript{\dag}, Yueting Zhuang\textsuperscript{\dag}, Guojie Song\textsuperscript{\S}\\
	\textsuperscript{\dag}College of Computer Science and Technology, Zhejiang University\\
	\textsuperscript{\ddag}State Grid Huzhou Power Supply Co. Ltd, China\\
	\textsuperscript{\S}Key Laboratory of Machine Perception, Ministry of Education, Peking University\\
}

\begin{document}

\maketitle
\input{abstract}
\input{intro}

\input{notations}
\input{model}

\input{exp}

\input{relate}
\input{conclude}
\pagebreak
\bibliographystyle{aaai}
\bibliography{reference}
\input{appendix}
\end{document}

%% file: abstract.tex
\begin{abstract}
Time series modeling
has attracted extensive research efforts;
however, achieving both reliable efficiency and interpretability from a unified model still remains a challenging problem. 
Among the literature, \textit{shapelets} 
offer interpretable and explanatory insights in the classification tasks, 
while most existing works 
ignore the differing representative power 
at different time slices, as well as (more importantly) the evolution pattern of shapelets. 
In this paper, 
we propose to 
extract \textit{time-aware shapelets} 
by designing a two-level timing factor. 
Moreover, 
we define and construct the \textit{shapelet evolution graph}, 
which captures how shapelets evolve over time and can be incorporated into the time series embeddings by graph embedding algorithms. 
To validate whether the representations obtained in this way can be applied effectively in various scenarios, 
we conduct experiments based on three public time series datasets, and two real-world datasets from different domains. 
Experimental results clearly show the improvements achieved by our approach compared with 
\numbaseline state-of-the-art baselines.
\end{abstract}

%% file: intro.tex
\section{Introduction}
\label{sec:intro}

Time series modeling aims to discover the temporal relationships within chronologically arranged data. 
The key issue here is how to extract the representative features of a time series. 
A large part of previous frameworks range from classical feature engineering and representation learning to deep learning based models. 
While these methods have achieved good performance~\cite{malhotra2016lstm,johnson2016composing}, they have also been subject to criticism regarding their lack of interpretability.  
On the other hand, \textit{shapelets}, the time series subsequences that are representative of a class~\cite{ye2011time}, can offer directly interpretable and explanatory insights in the classification scenario, 
and shapelet-based models have proven to be promising in various practical domains~\cite{ye2009time,xing2012early,lines2012shapelet,rakthanmanon2013fast,grabocka2014learning,hills2014classification,bostrom2017binary}.

Existing efforts have mainly considered shapelets as static. 
However, in the real world, 
shapelets are often dynamic, which is reflected in two respects. 
First, the same shapelet appearing at different time slices may have a range of different impacts. 
For instance, 
in the scenario of detecting electricity theft,  
low electricity consumption in summer or winter is more suspicious 
than it is in spring, 
as refrigeration or heating equipments costs more electrical power.
Second, determining the ways in which shapelets evolve is vital to a full understanding of a time series. 
In fact, shapelets with small values at a particular time can hardly distinguish an electricity thief from a normal user who indeed regularly consumes a low level of electricity. 
An alternative method would involve identifying users who once had high electricity consumption shapelets but suddenly consumes very few electrical power for a while. 
In other words, an important clue here is how shapelets evolve over time. 
We refer to the subsequences of a time series that are able to reflect its representativeness at different time slices as \textit{time-aware shapelets}. 

\begin{figure*}[ht!]
	\centering
	\includegraphics[width=0.9\textwidth]{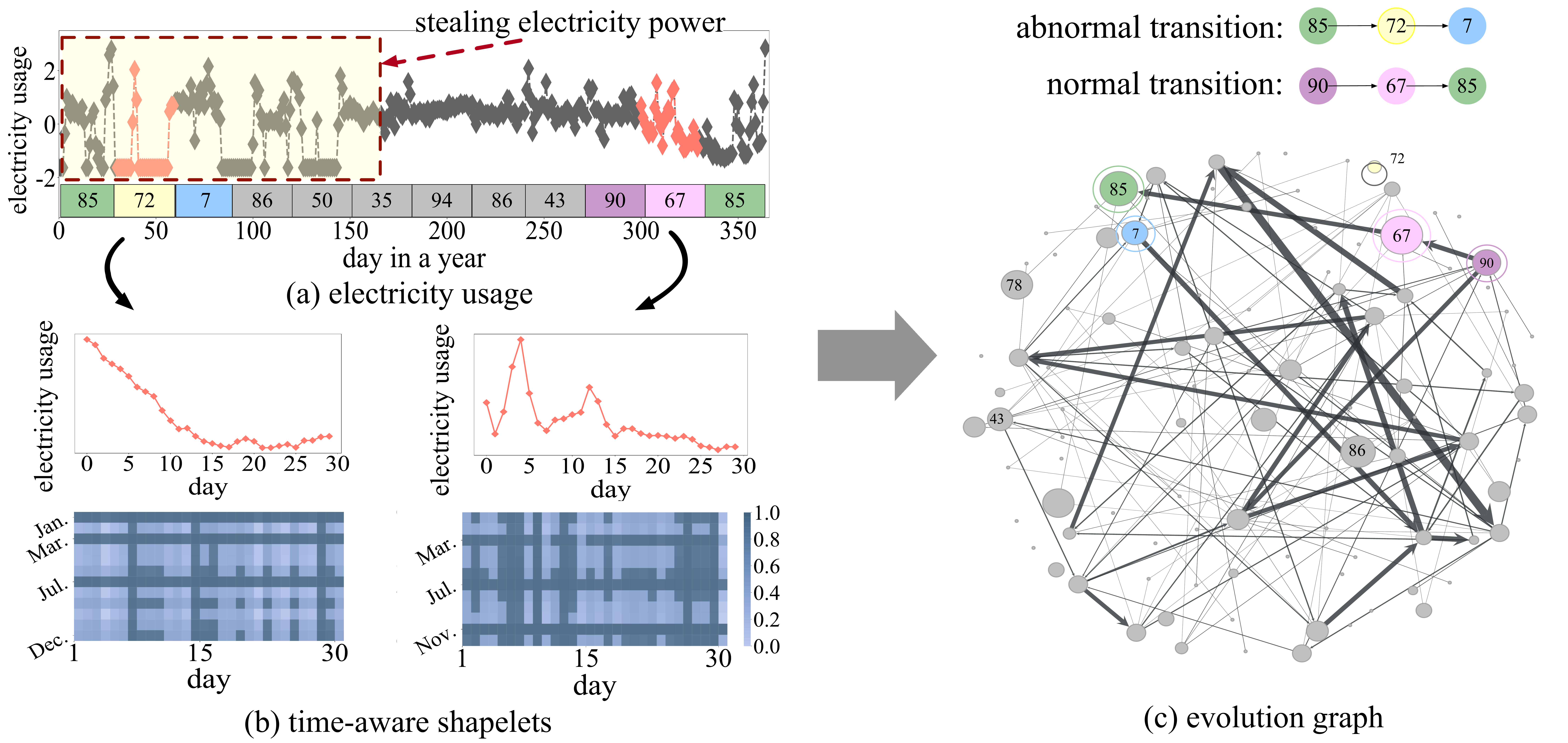}
	\caption{Illustration of (a) electricity consumption data, (b) two shapelets extracted from the observed time series and the corresponding timing factors, and (c) the shapelet evolution graph. 
		\small 
		In shapelet \textit{\#72}, there is an abnormal peak at the beginning, and then a continuous drop until the end, while its timing factor is highlighted mainly in Jan, Mar, and Jun. 
		As for the normal shapelet \textit{\#67}, weights in the timing factor may just reflect the frequency of its occurrence.
		In figure (c), the size of each vertex is proportional to its weighted in-degree, 
		and the width of the edge is proportional to its betweenness. 
		\normalsize 
	}
	\label{fig:graph:unigraph}
	\vspace{-2em}
\end{figure*}

There are several challenges involved in modeling the dynamics of shapelets. 
First, how can time-aware shapelets be defined, and then, be extracted? 
Traditional algorithms generate a set of static time series subsequences as candidates,
then select subsequences with the most discriminatory power
according to certain criterion~\cite{lines2012shapelet}. 
However, dynamic shapelets have not been well defined yet, 
and the criterion used before are often not differentiable if parameterized timing weights are added.
To the best of our knowledge, the question of how time-aware shapelets might be extracted remains open.
Second, how can the evolutions of shapelets be captured? 
One possible solution would be to explore the \textit{shapelet sequences}, which consist of ordered shapelets assigned to each \textit{segment} of the time series. 
However, as each segment may be assigned by several shapelets with different probabilities,  
enumerating all possible shapelet sequences is quite time-consuming,
where the time complexity is $\mathcal{O}(m^K)$ ($m$ is the number of segments and $K$ is the number of shapelets).
A more reliable solution would be to construct a transition matrix (graph) whereby each element (edge) represents the probability of a shapelet appearing after another,
but how to construct and analyze such matrix (graph) reasonably is nontrivial. 
Third, how can the evolution patterns of shapelets for modeling the given time series be utilized? 
Inspired by the recent success of representation learning~\cite{perozzi2014deepwalk,grover2016node2vec}, 
embedding the evolution patterns of shapelets into a latent feature space $\Sigma$ can be effective.
Then it seems feasible to represent the original time series by using a vector extended from $\Sigma$.
Designing a reasonable representation learning algorithm is the final challenge.

To address the abovementioned challenges, in this paper, we propose a novel approach to learn the representations of a time series by extracting time-aware shapelets and constructing a shapelet evolution graph. 
We first define a two-level timing factor to quantify the discriminatory power of shapelets at different time, 
then construct a graph to represent the evolution patterns of shapelets. 
\figref{fig:graph:unigraph} shows an example from real-world electricity consumption record data: 
\figref{fig:graph:unigraph}a demonstrates the one-year electricity usage of a user  who has stolen electrical power from January to May while using electrical power normally in the remaining months. 
We assign each month the most representative shapelet at that time and present the shapelets \textit{\#72} and \textit{\#67}, along with their timing factors in Figure~\ref{fig:graph:unigraph}b, where dark areas indicate that the corresponding shapelet is more discriminative relative to light areas. 
The shapelet evolution graph is presented in \figref{fig:graph:unigraph}c, illustrating how a shapelet would transfer from one to another \textit{in a normal case}:
for the normal electricity consumption record, there is a clear path for its shapelet transition (\textit{\#90} $\rightarrow$ \textit{\#67} $\rightarrow$ \textit{\#85}) in the graph. 
For the abnormal data, however, the path (\textit{\#85} $\rightarrow$ \textit{\#72} $\rightarrow$ \textit{\#7}) does not exist,
indicating that the connectivity of the shapelet transition path provides an evidential basis for detecting an abnormal time series. 
Finally, we translate the problem of learning representations of shapelets and time series into a graph embedding problem.  

We summarize our contributions to the field as follows:
1) We propose the concept of time-aware shapelets and design a learning algorithm to extract them;
2) We construct a shapelet evolution graph and translate the problem of representation learning for shapelets and time series into graph embedding;
and 3) We validate the effectiveness of our approach based on 
three public and two real-world datasets. 
Experimental results show that our approach achieves notably better performance when compared with \numbaseline state-of-the-art baselines.

%% file: notations.tex
\section{Preliminaries}
\label{sec:prelim}
A time series set $T=\{\bm{t}_1, \cdots, \bm{t}_{|T|}\}$, where each $\bm{t}$ contains $n$ chronologically arranged elements, 
i.e., $\bm{t}=\{\bm{x}_1, \cdots, \bm{x}_{n}\}$.
A \textit{segment} $\bm{s}$ of $\bm{t}$ is a contiguous subsequence, 
i.e., $\bm{s} = \{\bm{x}_i, \cdots, \bm{x}_j\}$. 
If $\bm{t}$ can be divided by $m$ segments of equal length $l$, then we have $\bm{t}=\{\{\bm{x}_{l*k+1}, \cdots, \bm{x}_{l*k + l}\}, 0\le k \le m-1\}$. 
To measure the dissimilarity of sequences, we denote the distance between two segments $\bm{s}_i$ and $\bm{s}_j$ as $d(\bm{s}_i, \bm{s}_j)$, where $d(\cdot, \cdot)$ can be intuitively formalized as the \emph{Euclidean Distance} (ED). 
But ED cannot deal with varied sequence length and timing shifts,
and in the context of time series modeling, 
time warping techniques are often used to address such problems.
The central idea of time warping is to find an appropriate alignment for the given pair of sequences,
where an alignment is defined as

\begin{definition}
	\textbf{Alignment.} 
	Given two segments $\bm{s}_i$ and $\bm{s}_j$ with length $l_i$ and $l_j$ respectively, 
	an alignment $\bm{a} = (\bm{a}_1, \bm{a}_2)$ is a pair of two index sequences of length $p$,
	satisfying that
	\begin{equation}
	\begin{split}
	1 \le \bm{a}_k(1) \le \cdots \le \bm{a}_k(p) = l_k, \\
	~\bm{a}_k(n+ 1) -  \bm{a}_k(n) \le 1, \\
	\mbox{for } k=i,j, \mbox{~and } 1 \le n \le p-1\\
	\end{split}
	\end{equation}
\end{definition}

We denote all possible alignments for two segments $\bm{s}_i$ and $\bm{s}_j$ as $\mathcal{A}(\bm{s}_i, \bm{s}_j)$,
then one popular time-warping based measurement, \emph{DTW (Dynamic Time Warping)}, can be illustrated as \eqnref{eq:dtw distance},
\noindent where $\tau(\bm{s}_i, \bm{s}_j | \bm{a})$ is a predefined dissimilarity for two sequences under the alignment $\bm{a}$~\cite{muller2007dynamic}.
We refer the alignment achieving the minimum in \eqnref{eq:dtw distance} as $\bm{a}^*$.
\begin{equation}
\label{eq:dtw distance}
d_{DTW}(\bm{s}_i, \bm{s}_j)  = \min\nolimits_{\bm{a}\in\mathcal{A}(\bm{s}_i, \bm{s}_j)}\tau(\bm{s}_i, \bm{s}_j | \bm{a}) 
\end{equation}

We can further measure the dissimilarity
between a segment $\bm{s}$ and a time series $\bm{t}=\{\bm{s}_1\cdots \bm{s}_{m}\}$.
Inspired by the literature that we often say a segment $\bm{s}$ is close to $\bm{t}$
if there exists some segment $\bm{s}'$ in $\bm{t}$  between which the distance of $\bm{s}$ is rather small, 
we define the distance between $\bm{s}$ and $\bm{t}$ as
\begin{equation}
D(\bm{s}, \bm{t})  = \min\nolimits_{1\le k \le m}d(\bm{s}, \bm{s}_k)
\label{eq:s-t distance}
\end{equation}

Based on these definitions,
previous work have proposed novel methods to extract typical subsequences, i.e., shapelets, 
to distinguish the \textit{representative power} of segments: 
\begin{definition}
	\textbf{Shapelet.} 
	A shapelet $\bm{v}$ is a segment that is representative of a certain class. 
	More precisely, it can separate $T$ into two smaller sets, 
	one that is close to $\bm{v}$ and another far from $\bm{v}$ by some specific criteria, 
	such that for a time series classification task, positive and negative samples can be put into different groups. 
	The criteria can be formalized as
	
	\begin{equation}
	\label{eq:raw-loss}
	\mathcal{L} = -g(S_{pos}(\bm{v}, T), S_{neg}(\bm{v}, T))
	\end{equation}
\end{definition}

$\mathcal{L}$ measures the dissimilarity between positive and negative samples towards the shapelet $\bm{v}$.
$S_*(\bm{v}, T)$ denotes the set of distances with respect to a specific group $T_*$, i.e., positive or negative class;
the function $g$ takes two finite sets as input,  returns a scalar value to indicate how far these two sets are,
and it can be \textit{information gain}~\cite{ye2011time}, or some dissimilarity measurements on sets, i.e., \textit{KL} divergence.

%% file: model.tex
\section{Time2Graph Framework}
\label{sec:model}
In this section, we present a novel representation learning algorithm for time series modeling. 
We name the proposed framework as \textit{Time2Graph}, as it transforms time series to a graph with shapelets and their transitions. 
We extract time-aware shapelets from a large pool of candidates (\secref{subsec:shapelet}),
then construct the \textit{Shapelet Evolution Graph} to capture the correlations between shapelets (\secref{subsec:graph}), 
and finally learn the time series representation vectors (\secref{subsec:representation}) by concatenating segment embeddings which is composed by shapelet embeddings obtained from the \textit{Shapelet Evolution Graph}.

\subsection{Time-Aware Shapelet Extraction}
\label{subsec:shapelet}
The traditional definition of shapelets ignores that subsequences may have different representative powers at different time. 
For example, low consumption of electrical power in spring is normal, whereas it is a strong signal for identifying abnormal users in summer, when high temperatures often lead to high electricity usage. 
Therefore, we consider time-aware shapelets in this paper. 

We define two factors for quantitatively measuring the timing effects of shapelets at different levels. 
Specifically, we introduce the \textit{local factor}  $\bm{w_n}$ to denote the inner importance of the $n$-th element of a particular shapelet,
then the distance between a shapelet $\bm{v}$ and a segment $\bm{s}$ is redefined as 

\begin{equation}
\label{eq:w-seg-distance}
\begin{split}
\hat{d}(\bm{v}, \bm{s}|\bm{w}) &= \tau(\bm{v}, \bm{s} | \bm{a}^*, \bm{w}) \\
&= (\sum\nolimits_{k=1}^{p}\bm{w}_{\bm{a}^*_1(k)} \cdot (\bm{v}_{a^*_1(k)} - \bm{s}_{\bm{a}^*_2(k)})^2)^{\frac{1}{2}}
\end{split}
\end{equation}
where $\bm{a}^*$ refers to the best alignment for \emph{DTW} which we have discussed in \eqnref{eq:dtw distance}.
The intuitive explanation of \eqnref{eq:w-seg-distance} is to project the weight $\bm{w}$ onto the \emph{DTW} alignment path.
On the other hand, at a \textit{global level}, we aim to measure the timing effects across segments on the discriminatory power of shapelets. 
It is inspired from the intuition that shapelets may represent totally different meaning at different time steps,
and it is straightforward to measure such deviations by adding segment-level weights. 
Formally, we set a \textit{global factor} $\bm{u}_m$ to capture the cross-segments influence, 
then the distance between a shapelet $\bm{v}$ and a time series $\bm{t}$ can be rewritten as 

\begin{equation}
\label{eq:s-t timing distance}
\hat{D}(\bm{v}, \bm{t} | \bm{w}, \bm{u}) = \min\limits_{1\le k \le m} \bm{u}_k \cdot \hat{d}(\bm{v}, \bm{s}_k | \bm{w})
\end{equation}
where $\bm{t}$ is divided into $m$ segments, i.e., $\bm{t} = \{\bm{s}_1, \cdots, \bm{s}_m\}$.
\eqnref{eq:s-t timing distance} denotes the two-level time-aware distance between a shapelet $\bm{v}$ and a time series $\bm{t}$,
and the parameters $\bm{w}, \bm{u}$ associated with each specific shapelet can be learned separately under some proper criteria. 
Given a classification task, we establish a supervised learning method to select
the most important time-aware shapelets and learn corresponding timing factors $\bm{w}_i$ and $\bm{u}_i$ for each shapelet $\bm{v}_i$. 
In particular, we have a pool of segments as shapelet candidates that generated by \algref{alg:candidate} (see in \secref{subsec:appendix:alg}), and a set of time series $T$ with labels. 
We only consider binary classification here, to which it is straightforward to extend multinomial classification. 
For each shapelet candidate $\bm{v}$, we modify \eqnref{eq:raw-loss} as 
\begin{equation}
\label{eq:timing-loss}
\hat{\mathcal{L}} =
- g(S_{pos}(\bm{v}, T), S_{neg}(\bm{v}, T)) + \lambda ||\bm{w}|| + \epsilon ||\bm{u}||
\end{equation}
\noindent where 
$\lambda$ and $\epsilon$ are the hyperparameters of penalties, and as mentioned above, 
the differentiable function $g(\cdot, \cdot)$ measures the distances between distributions of two finite sets.
In practice, we assume that the given sets both follow some particular distributions, e.g., Gaussian Distribution. 
We can easily estimate the distribution parameters by closed-form solutions,
and then the gradients of $g$ can be derived from those differentiable parameters.
After learning the timing factors from shapelet candidates, 
we select the top $K$ shapelets with minimal loss in \eqnref{eq:timing-loss}.  
The procedure for extracting time-aware shapelets is formalized in \algref{alg:timing factor} in \secref{subsec:appendix:alg}.

\subsection{Shapelet Evolution Graph}
\label{subsec:graph}
After obtaining shapelets, many works use BoP~\cite{baydogan2013bag} or similar methods to represent the time series, 
but these algorithms ignore the correlations between shapelets.  
Here, \textit{correlations} consist of the co-occurrence along with the occurrence order.
To capture such relationship,
we propose the concept of 
\textit{Shapelet Evolution Graph} as follows: 
\begin{definition}
\textbf{Shapelet Evolution Graph.} It is a directed and weighted graph $G=(V, E)$ in which $V$ consists of $K$ vertices, each denoting a shapelet,
and each directed edge $e_{ij} \in E$ is associated with a weight $w_{ij}$, 
indicating the occurrence probability of shapelet $\bm{v}_i \in V$ followed by another shapelet $\bm{v}_j \in V$ in the same time series. 
\end{definition} 

\vpara{Graph Construction.} 
We first assign each segment $\bm{s}_i$ of each time series
to several shapelets that have the closest distances to $\bm{s}_i$
according to the time-aware dissimilarity.
Then a problem naturally rises as how far the distance would be considered as \textit{closest}?
One simple but effective solution is to predefine a threshold $\delta$ such that distances less than $\delta$ are treated as close,
and in practice, we can determine $\delta$ by experimental statistics on the training dataset.
For convenience, we denote those shapelets assigned to segment $\bm{s}_i$ as $\bm{v}_{i,*}$, 
and say that $\bm{v}_{ij}$ is the $j$-th assignment of $\bm{s}_i$.
For the purpose of measuring how reasonable our assignment is,
we standardize assignment probability $\bm{p}_{i,j}$ as 
\begin{equation}
\label{eq:norm dist}
\bm{p}_{i, j} = \frac{
	\max(\hat{d_{i,*}}(\bm{v}_{i, *}, \bm{s}_i)) - \hat{d_{i,j}}(\bm{v}_{i, j}, \bm{s}_i)
}{
	\max(\hat{d_{i,*}}(\bm{v}_{i, *}, \bm{s}_i)) - \min(\hat{d_{i,*}}(\bm{v}_{i, *}, \bm{s}_i))
}
\end{equation}

\noindent where 
$\hat{d_{i,*}}(\bm{v}_{i, *}, \bm{s}_i) = \bm{u}_*[i] * \hat{d}(\bm{v}_{i, *}, \bm{s}_i |\bm{w}_*)$ (\eqnref{eq:w-seg-distance}),
with the constraint that $\hat{d_{i, *}} \le \delta$. 
So the shapelets set $\bm{v}_{i,*}$ is assigned to segment $\bm{s}_i$ with probability $\bm{p}_{i,*}$,  
and $\bm{v}_{i+1,*}$ is assigned to $\bm{s}_{i+1}$ with probability $\bm{p}_{i+1, *}$ respectively,
where $\bm{s}_{i}$ is followed by $\bm{s}_{i+1}$ in one time series. 
Then, for each pair $(j, k)$, we create a weighted edge from shapelet $\bm{v}_{i, j}$ to $\bm{v}_{i+1, k}$
with weight $\bm{p}_{i,j} \cdot \bm{p}_{i+1,k}$. 
and merge all duplicated edges as one by summing up their weights.
Finally,  we normalize the edge weights sourced from each node as 1,
which naturally interprets the edge weight between each pair of nodes, i.e., $\bm{v}_i$ and $\bm{v}_j$ into the conditional probability $P(\bm{v}_j | \bm{v}_i)$
that shapelet $\bm{v}_i$ being transformed into $\bm{v}_j$ in an adjacent time step.
See \algref{alg:graph construction} for an overview of graph construction procedure.
\begin{algorithm}[h]
	\caption{Shapelet Evolution Graph Construction}
	\label{alg:graph construction}		
	\begin{algorithmic}[1]
		\REQUIRE ~~\\
		time series set $T = \{\bm{t}_1, \cdots, \bm{t}_{|T|}\}$,\\
		$K$ shapelets $\{\bm{v}_1\cdots\bm{v}_K\}$, distance threshold $\delta$
		\ENSURE ~~\\
		Shapelet evolution graph $G$\\
		\STATE Initialize the graph $G$ with $K$ vertices
		\FORALL{segment $\bm{s}_i$ of $\bm{t}$ in $T$}
		\label{alg:gc:assign:begin}
		\FORALL{shapelet $\bm{v}_j$ where $\hat{d}(\bm{v}_j, \bm{s}|\bm{w}_j) * \bm{u}_j[i] \le \delta$}
		\STATE  Assign $\bm{v}_j$ to $\bm{s}_i$ with probability defined in \eqnref{eq:norm dist}
		\ENDFOR
		\ENDFOR
		\label{alg:gc:assign:end}
		
		\FORALL{adjacent segment pair $(\bm{s}_i, \bm{s}_{i+1})$ of $\bm{t}$ in $T$}
		\FORALL{assigned shapelet pair $(\bm{v}_{i, j}, \bm{v}_{i+1, k})$}
		\STATE  Add directed edge $e_{j, k}$ with weight $\bm{p}_{i, j} * \bm{p}_{i+1, k}$
		\ENDFOR
		\ENDFOR
		\STATE Normalize the edge weights for each vertex 
		\RETURN $G$
	\end{algorithmic}
\end{algorithm}

\subsection{Representation Learning}
\label{subsec:representation}
Finally, we learn the representations for both the shapelets and the given time series by using the shapelet evolution graph constructed as above. 
We first employ an existing graph embedding algorithm (DeepWalk~\cite{perozzi2014deepwalk}) to obtain vertex (shapelet) representation vectors $\bm{\mu} \in \mathbb{R}^B$, 
where $B$ is the embedding size (latent dimension). 
In our case, a path in $G$ intuitively reflects possible transitions between shapelets. 

Next, for a  time series $\bm{t} = \{\bm{s}_1\cdots \bm{s}_{m}\}$ with corresponding assigned shapelets $\{\bm{v}_{1, *}\cdots\bm{v}_{m, *}\}$ and assignment probabilities  
$\{\bm{p}_{1, *},$ $\cdots\bm{p}_{m, *}\}$, 
we retrieve each shapelet $\bm{v}_{i, j}$'s representation vector $\bm{\mu}(\bm{v}_{i,j})$ multiplied by assignment probability $\bm{p}_{i, j}$, and sum them over each segment.
If there exists some segment that we cannot assign any shapelets to it by applying the predefined distance threshold, the embeddings of this segment would be left as empty.
It is reasonable since shapelet embeddings are always non-zero which is guaranteed by graph embedding models 
(more precisely, shapelet embeddings are bound to be normalized),
so segments without shapelet assignments are clearly distinguished by empty values.
By far, we get the segment representations, and finally concatenate all those $m$ segment embedding vectors to obtain the representation vector $\bm{\Phi}$ for time series $\bm{t}$ as follows: 

\begin{equation}
\label{eq:representation}
\bm{\Phi}_i=(\sum\limits_{j}\bm{p}_{i,j}\cdot\bm{\mu}(\bm{v}_{i,j})), \ 1 \le i \le m
\end{equation}

The representation vector of the time series can then be applied as features for various time series classification tasks,
by the way of feeding the embedding features into an outer classifier.
See the formulation of the representation learning framework in \algref{alg:timeseries-embedding},
and the details of the down-streaming classification tasks are introduced in \secref{subsec:exp setup}.

	\begin{algorithm}[h]
	\caption{Time Series Embedding Framework}
	\label{alg:timeseries-embedding}		
	\begin{algorithmic}[1]
		\REQUIRE ~~\\
		a time series $\bm{t} = \{\bm{s}_1\cdots\bm{s}_{m}\}$,\\
		$K$ time-aware shapelets $\{\bm{v}_1\cdots\bm{v}_K\}$,\\
		shapelet evolution graph $G$, graph embedding size $B$
		\ENSURE ~~\\
		time series embedding vector $\bm{\Phi}$\\
		\label{alg4:output}
		\STATE Embeds graph $G$ as $\bm{\mu}^B$ by \emph{deepwalk} with\\
		\emph{representation-size} as $B$
		\STATE Initialize $\bm{\Phi}$ as an empty vector.
		\label{alg4:init}
		\FORALL{segment $\bm{s}_i$ in $\bm{t}, 1\le i \le m$}
		\STATE $temp$ $\leftarrow$ zero vectors with size $B$
		\FORALL{shapelet $\bm{v}_{i, j}$ that assigned to $\bm{s}_i$}
		\STATE $temp$ += $\bm{p}_{i, j} * \bm{\mu}^B(\bm{v}_{i, j})$, where shapelet assignment can be referred from 
		\algref{alg:graph construction}
		\ENDFOR
		\STATE $\bm{\Phi}$.append($temp$)
		\ENDFOR
		\RETURN $\bm{\Phi}$
	\end{algorithmic}
\end{algorithm}

%% file: exp.tex
\section{Experiments}
\label{sec:exp}

\subsection{Experimental Setup}
\label{subsec:exp setup}
We use three public datasets, \textit{Earthquakes} (EQS), \textit{WormsTwoClass} (WTC) and \textit{Strawberry} (STB) from the \textit{UCR Time Series Archive}~\cite{UCRArchive2018}, 
along with two real-world datasets, Electricity Consumption Records (\textit{\sgdata}) from State Grid of China and Network Traffic Flow (\textit{\teledata}) from \textit{China Telecom.}, to validate our proposed model. 
Table~\ref{tb:exp:dataset} shows the overall statistics of those five datasets:
\begin{table}[h]
	\centering
	\captionsetup{justification=centering}
	\addtolength{\tabcolsep}{-1pt}
	\begin{tabular}{l|ccccc}
		\toprule
		\textbf{\diagbox[width=2.4cm]{Metric}{Dataset}} & \multirow{1}{*}{\textbf{\ucreqk}} & \multirow{1}{*}{\textbf{\ucrworms}}  
		&\multirow{1}{*}{\textbf{\ucrstra}} & \multirow{1}{*}{\textbf{\sgdata}}  &\multirow{1}{*}{\textbf{\teledata}} \\
		\midrule
		\#(time series) & 461 & 258 & 983 & 60,872& 5,950 \\
		positive ratio(\%) & 25.3 & 42.2 & 64.3  & 2.3 & 6.4 \\
		\bottomrule
	\end{tabular}
	\normalsize
	
	\caption{
		Overall statistics of 5 datasets in the experiments.
	}
	\label{tb:exp:dataset}
	\vspace{-1em}
\end{table}

\begin{table*}[t]
	\centering
	\captionsetup{justification=centering}
	\addtolength{\tabcolsep}{1pt}
	\begin{tabular}{c|ccc|p{0.8cm}p{0.8cm}p{0.8cm}|p{0.8cm}p{0.8cm}p{0.8cm}}
		\toprule
		& \multicolumn{3}{c|}{\textbf{public dataset}} & \multicolumn{6}{c}{\textbf{real-world dataset}} \\
		\hline
		\multirow{2}{*}{\textbf{\diagbox{Methods}{Datasets}}} & \multirow{1}{*}{\textbf{\ucreqk}} & \multirow{1}{*}{\textbf{\ucrworms}}  &\multirow{1}{*}{\textbf{\ucrstra}} & \multicolumn{3}{c|}{\textbf{\sgdata}} & \multicolumn{3}{c}{\textbf{\teledata}} \\
		
		& \multicolumn{3}{c|}{Accuracy}& Prec & Recall &$F_1$ & Prec & Recall &$F_1$ \\
		\midrule
		
		NN-ED & 68.22 & 62.41& 95.60  & 18.71 & 10.48 & 13.44 & 37.71 & 46.35 & 41.59 \\
		NN-DTW & 70.31 & 68.16 & 95.53  & 15.52 & 18.15 & 16.73 & 33.20 & 43.75 & 37.75 \\
		NN-WDTW & 69.50 & 67.74 & 95.44  & 15.52 & 18.15 & 16.73 & 35.29 & 46.86 & 40.27 \\
		NN-CID & 69.41 & 69.56 & 95.51 &  18.18 & 13.71 & 15.63 & 32.56 & 43.75 & 37.33 \\
		DDDTW & 70.79 & 70.92 & 95.60  & 18.78 & 13.71 & 15.85 & 30.48 & 42.71 & 35.58 \\
		
		\midrule
		\textit{XGBoost} (origin) & 74.82 & 62.34 & 95.92  & 38.36 & 19.48 & 25.86 & 71.43 & 17.86 & 28.57 \\
		\textit{XGBoost} (feature) & 75.54 & 64.94 & 97.03*  & 56.82 & 16.23 & 25.25 & 80.00* & 21.43 & 33.80 \\
		BoP & 74.80 & 74.42* & 96.45 &  14.86 & 4.44 & 6.83 & 43.40 & 47.92 & 45.55 \\
		TSF & 74.67 & 68.51 & 96.27 &  26.32 & 2.02 & 3.75 & 57.52 & 33.85 & 42.62 \\
		EE & 73.50 & 71.74 & 95.88  & 10.18 & 33.47 & 15.62 & 42.98 & 27.08 & 33.23 \\
		SAXVSM & 73.76 & 72.10 & 96.97 & 21.59 & 42.74 & 28.69 & 30.19 & 50.00 & 37.65\\
		
		\midrule
		LS & 74.22 & 73.57 & 92.49 &  0.00 & 0.00 & 0.00 & 0.00 & 0.00 & 0.00 \\
		FS & 74.66 & 70.58 & 91.66 &  10.45 & 79.84* & 18.48 & 63.55 & 35.42 & 45.49 \\
		LPS & 66.78 & 74.26 & 96.35 & 17.00 & 24.19 & 19.97 & 24.17 & 30.21 & 26.85 \\
		
		\midrule
		MLP & 70.29  & 59.86  & 96.58   & 0.00 & 0.00 & 0.00 & 0.00 & 0.00 & 0.00 \\
		LSTM & 74.82  & 42.86  & 63.84  & 13.64 & 31.86 & 19.11 &  7.22 & 16.67 & 10.08 \\
		VAE & 71.22  & 62.34  & 71.35  & 19.02 & 14.11 & 16.20 &  59.79 & 30.21 & 40.14 \\
		
		\midrule
		Shapelet-Seq & 75.53 & 55.84 & 78.10  & 14.37 & 66.94& 23.66 & 18.45 & 61.98* & 28.44 \\
		\modelname-static & 76.98 & 70.13  & 95.68  & 
		33.81 & 29.22 & 31.36 & 80.00* & 28.57 & 42.11 \\
		\textbf{\modelname} & \textbf{79.14}* & \textbf{72.73}  & \textbf{96.76} & 
		\textbf{30.10*} & \textbf{40.26} & \textbf{34.44}* & \textbf{71.52}* & \textbf{56.25} & \textbf{62.97}*\\
		\bottomrule
	\end{tabular}
	\normalsize
	\caption{
		Comparison of classification performance on the public and real-world datasets (\%). 
		\small Results of the \modelname~model are bold, and a star (*) means the best performance among all methods. 
	}
	\label{tb:exp:performance}
	\vspace{-1em}
\end{table*} 

The description of three public datasets can be found on the \textit{UCR Archive}
and referred in \secref{subsec:appendix:dataset}. 
Here we briefly introduce the two real-world datasets as follows:

\vpara{Electricity Consumption Records (\sgdata).} 
This dataset is provided by the State Grid Corporation of China
and contains the daily electricity consumption records (K$\cdot$Wh) of 60,872 users over the span of one year (2017). 
For every user, it records the daily total electricity usage, on-peak usage, and off-peak usage. 
Some users may take unauthorized actions concerning the electricity meter or power supply lines to cut costs (i.e., electricity theft),
and there are a total of 1,433 (2.3\%) users who have been manually confirmed as having stolen electrical power. 
Given users and their electricity consumption record, the task is to determine which users have stolen electrical power in the past year.

\vpara{Network Traffic Flow (\teledata).}  
This dataset is provided by China Telecom, the major mobile telecommunications service provider in China.
It consists of 5,950 network traffic series, each of which describes the hourly inflow and outflow of different servers, from April 6th 2017 to May 15th 2017.
When an abnormal flow goes through server ports and some process is suddenly dead, 
an alarm state is recorded by the operating system (objective ground-truth);
there are 383 (6.4\%) servers with abnormal flow series.
The goal is to use the daily network traffic data to detect whether there are abnormal flows during a period. 

We compare our proposed \textit{\modelname} model with several groups of the state-of-the-art baselines:

\vpara{Distance-based Models.}
	Previous work has stated that in most time series classification tasks, 1-NN-based methods are hard to beat~\cite{wang2013experimental,Bagnall2017}. 
	As for the distance metric applied in 1-NN, we use Euclidean Distance (ED), Dynamic Time Warping (DTW), Weighed DTW (WDTW)~\cite{jeong2011weighted}, Complexity-Invariant Distance (CID)~\cite{batista2014cid} and Derivative DTW (DDDTW)~\cite{gorecki2013using} as candidates. 
	
\vpara{Feature-based Models.}
	We first extract some statistical features (the mean, standard deviation, etc.),  or just take the raw time series as input (\textit{XGBoost} (origin/feature)),
	and use the same outer classifier as which \modelname uses (\textit{xgboost}) 
	to validate the effectiveness of the representation learning framework.
	Besides, several popular feature-based algorithms have been proposed for time series classification tasks.
	In this paper, we choose some typical algorithms to compare with our model:
	Bag-of-Patterns (BoP)~\cite{lin2012rotation}, Time Series Forest (TSF)~\cite{deng2013time}, 
	Elastic Ensembles (EE)~\cite{lines2015time}, 
	and Vector Space Model using SAX (SAXVSM)~\cite{senin2013sax}.

\vpara{Shapelet-based Models.}
	Another typical group of algorithms that are highly-related with our proposed model extracts \textit{shapelets} to 
	capture the intrinsic features of the original time series data. 
	In this paper, we use several famous shapelet-based frameworks as baselines:
	Learn Time Series Shapelets (LS)~\cite{grabocka2014learning},
	Fast Shapelets (FS)~\cite{rakthanmanon2013fast} and 
	Learned Pattern Similarity (LPS)~\cite{baydogan2016time}. 
	
\vpara{Deep learning models.}
	We consider three commonly-used deep models, MLP, LSTM and VAE, 
	due to their efficacy in feature-representation tasks and processing time series data. 
	
\vpara{\textit{\modelname}~variants.}
	We also compare \modelname~model with its derivatives by modifying some key components to see how they fare: 
	a) We sample the most possible shapelet sequence (i.e., each segment is assigned with highest assignment probability) for each time series, 
	and use \textit{LSTM} to conduct end-to-end classifications, denoted as \textit{Shapelet-Seq}; 
	b) We learn shapelets without considering timing factors, and embed them in the same way of \modelname, 
	and refer this method as \modelname-static. 
		
We choose \textit{XGBoost}~\cite{chen2016xgboost} as the outer classifier, 
and use 5-fold nested cross-validation to conduct fine-tuning on hyper-parameters. 
Explanations for the choice of the graph embedding algorithm and the outer classifier (\ref{subsec:implement}), 
along with implementation details and parameter settings (\ref{subsec:appendix-para}) are listed
in the supplementary material.

\subsection{Comparison Results}
\label{subsec:exp:result}

\tableref{tb:exp:performance} shows the comparison results for classification tasks. 
All three public datasets from \textit{UCR Archive} uniformly use accuracy as evaluation metric,
and for those two real-world datasets, which are both very imbalanced, 
we show the prediction precision, recall and F1 score.

We conclude from \tableref{tb:exp:performance} that the \modelname~model achieves competitive performance on the three public datasets. 
Specifically, on the \textit{EQS} dataset,  \modelname
 achieves the highest accuracy (79.14\%),
while on the other two datasets, \modelname~also beats most of the baseline methods.
When it comes to the real-world datasets, 
it is clear that \modelname significantly outperforms all baselines in the F1 score metric.
Even though some baseline methods achieve higher precision or recall, all of them seem to encounter biases on positive or negative samples.

We next compare \modelname with its variance. As mentioned in \secref{sec:intro},
\textit{Shapelet-Seq} model suffers from the size of possible sequences,
and when we only sample sequences with the highest probability, 
its performance fails to match several baselines,
since a substantial amount of information is lost during sequence sub-sampling.
The performance incrementation from \modelname-Static to \modelname~ demonstrates the predictive power brought by time-aware shapelets, and the additional interpretability and insights derived from timing factors are shown in \secref{subsec:case:shapelet}.
In summary, the \modelname~model is better at finding effective patterns,
as well as capturing evolutionary characteristics in time series.
\subsection{Parameter Analysis}
\label{subsec:parameter}
\begin{figure}[h!]
	\begin{minipage}{0.47\textwidth}
		\centering
		\begin{subfigure}[t]{0.31\textwidth}
			\includegraphics[width=\textwidth]{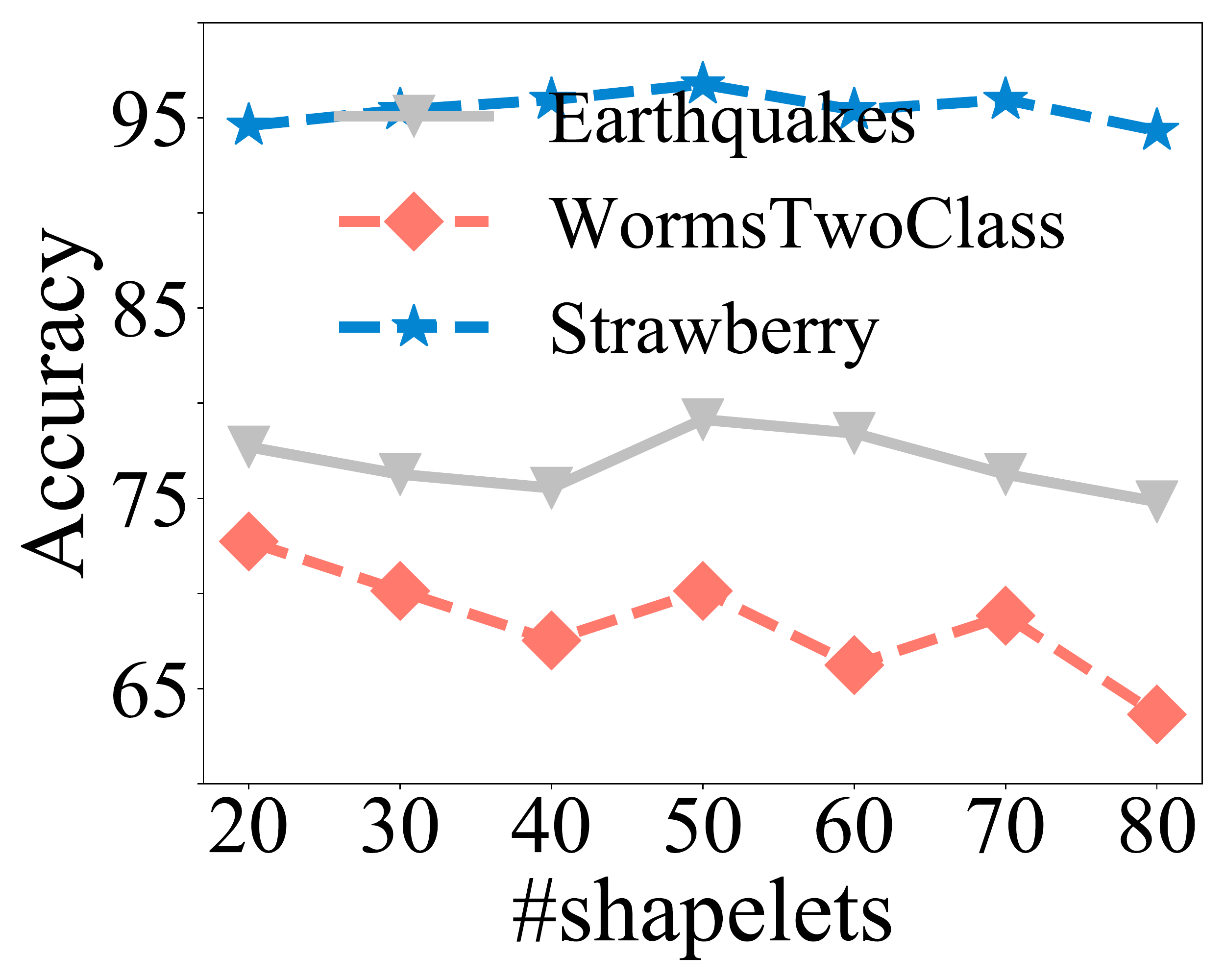}
			\caption{\#shapelets}
			\label{fig:predict:hyperK}
		\end{subfigure}
		\hfill
		\begin{subfigure}[t]{0.31\textwidth}
			\includegraphics[width=\textwidth]{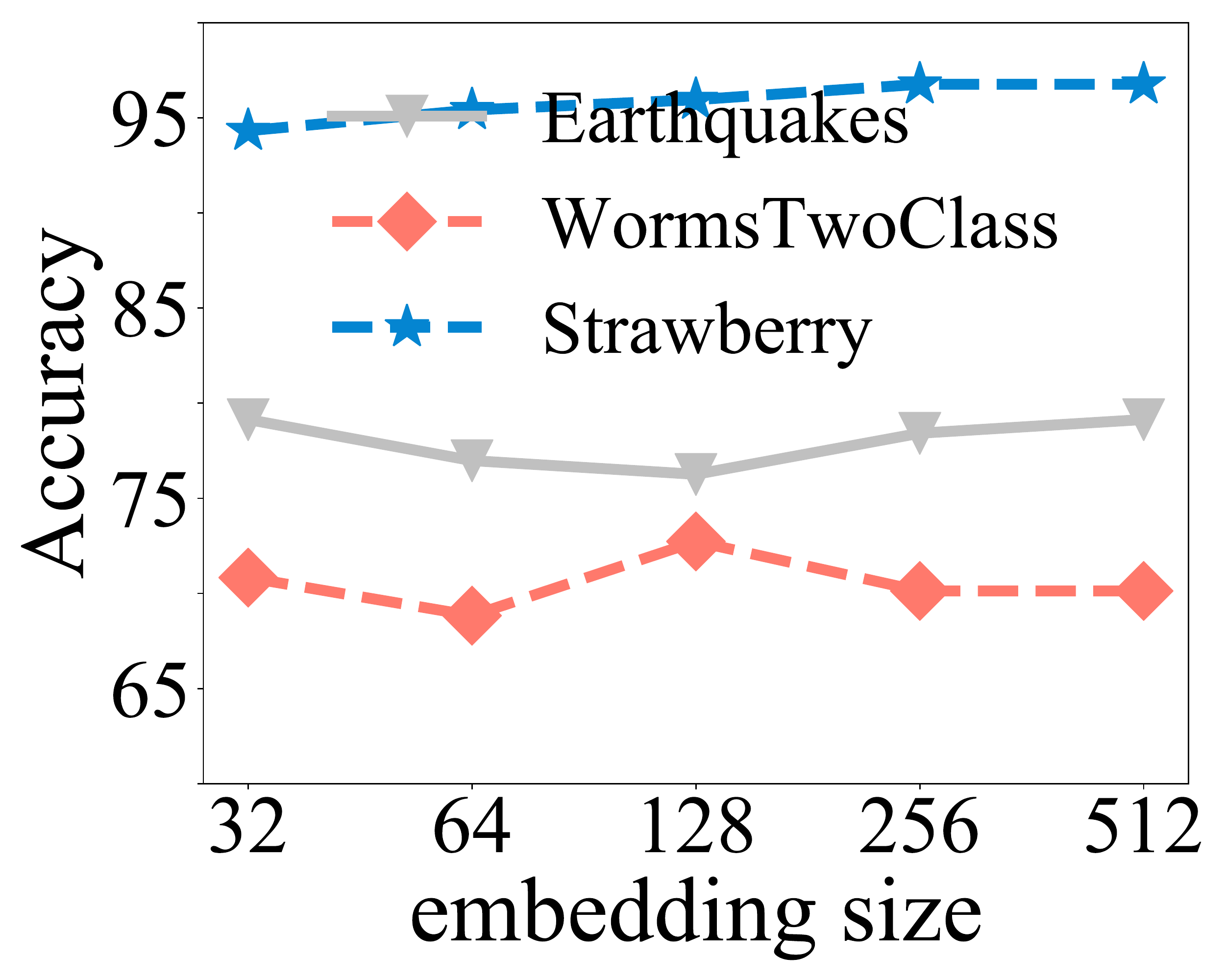}
			\caption{embed size}
			\label{fig:predict:hyper-size}
		\end{subfigure}
		\hfill
		\begin{subfigure}[t]{0.31\textwidth}
			\includegraphics[width=\textwidth]{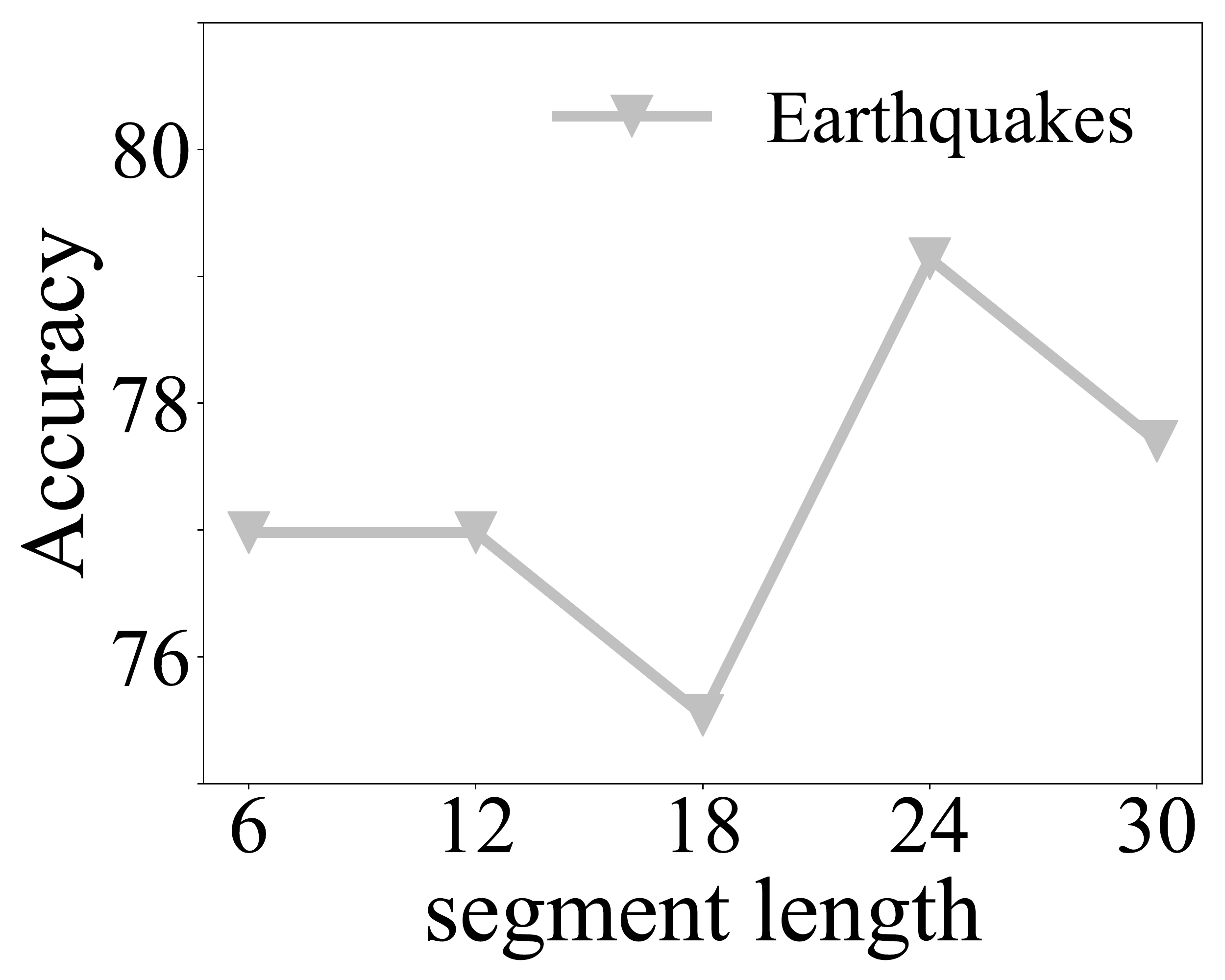}
			\caption{seg-length}
			\label{fig:predict:hyper-seg}
		\end{subfigure}
		\caption{
			Parameter analysis. 
		}
		\label{fig:predict:hyper}
		\normalsize
		\vspace{-2em}
	\end{minipage}
\end{figure}
We examine the sensitivities of three important hyperparameters:
number of selected shapelets $K$, graph embedding size $B$ and segment length $l$. 
Due to space limitations, 
we only present the results for the public datasets, which are shown in \figref{fig:predict:hyper}.
From the results, we see 
that $K$ should be large enough to capture a sufficient number of shapelets; 
while when $K$ is too large, it will bring in less representative shapelets as noise 
(\figref{fig:predict:hyperK}). 
Another parameter that should be tuned is the segment length $l$. 
We can see from \figref{fig:predict:hyper-seg} that 
it achieves the best results when $l$ is 24, which is exactly the same as the number of hours in a day.
It seems not to be a coincidence that, in \sgdata dataset, the best segment length is 30, i.e., the number of days in a month, 
while the optimal choice for \teledata is 24, again the number of hours in a day.
We may conclude that the segment length $l$ should be carefully selected based on the physical characteristics of the original data source.
As for the embedding size, we see that accuracy improves in most cases when it is increasing (\figref{fig:predict:hyper}b), whereas the efficiency will be sacrificed. 
It is also difficult to train the outer classifiers for features with dimensions that are too large; 
accordingly, an appropriate graph embedding size is necessary for better performance.

\begin{figure*}[h!]
	\centering
	\includegraphics[width=0.9\textwidth]{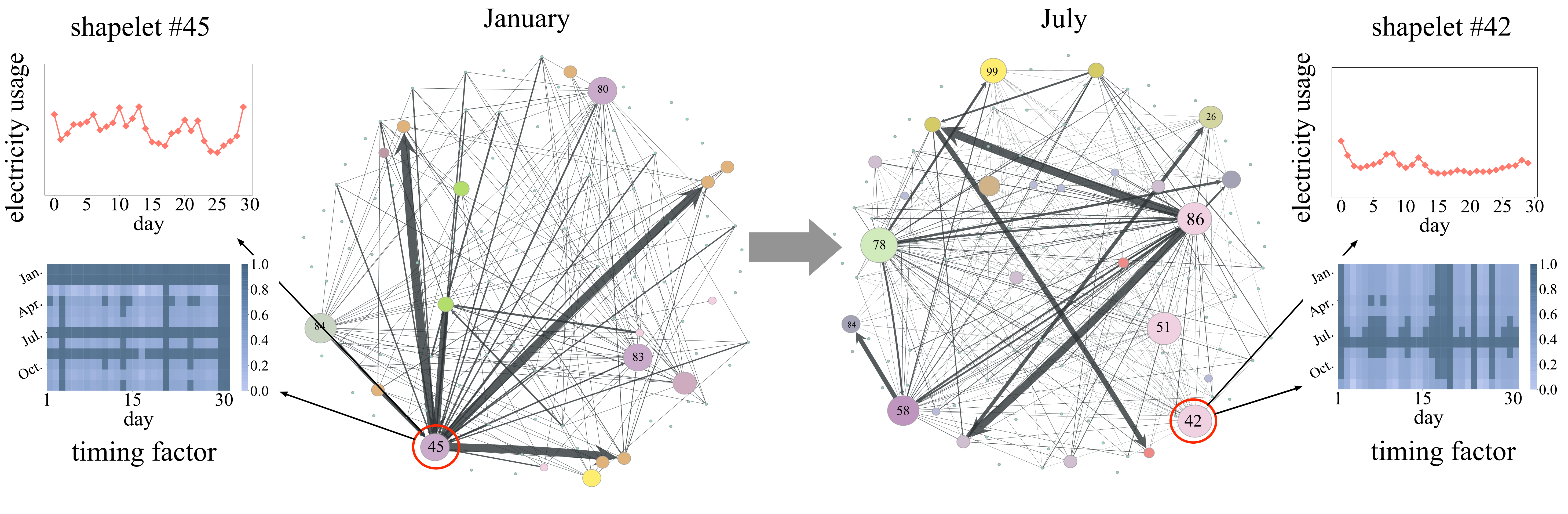}
	\caption{Shapelet evolution graphs at different times. 
		\small
		The positions of vertices in the two graphs are the same.  
		The vertex size 
		is proportional to its weighted in-degree, the same as the edge width to its betweenness.
		\normalsize
	}
	\label{fig:graph:multi-graph}
	\vspace{-1em}
\end{figure*}

\begin{figure}[ht!]
	\begin{minipage}{0.47\textwidth}
		\centering
		\begin{subfigure}[t]{\textwidth}
			\includegraphics[width=\textwidth]{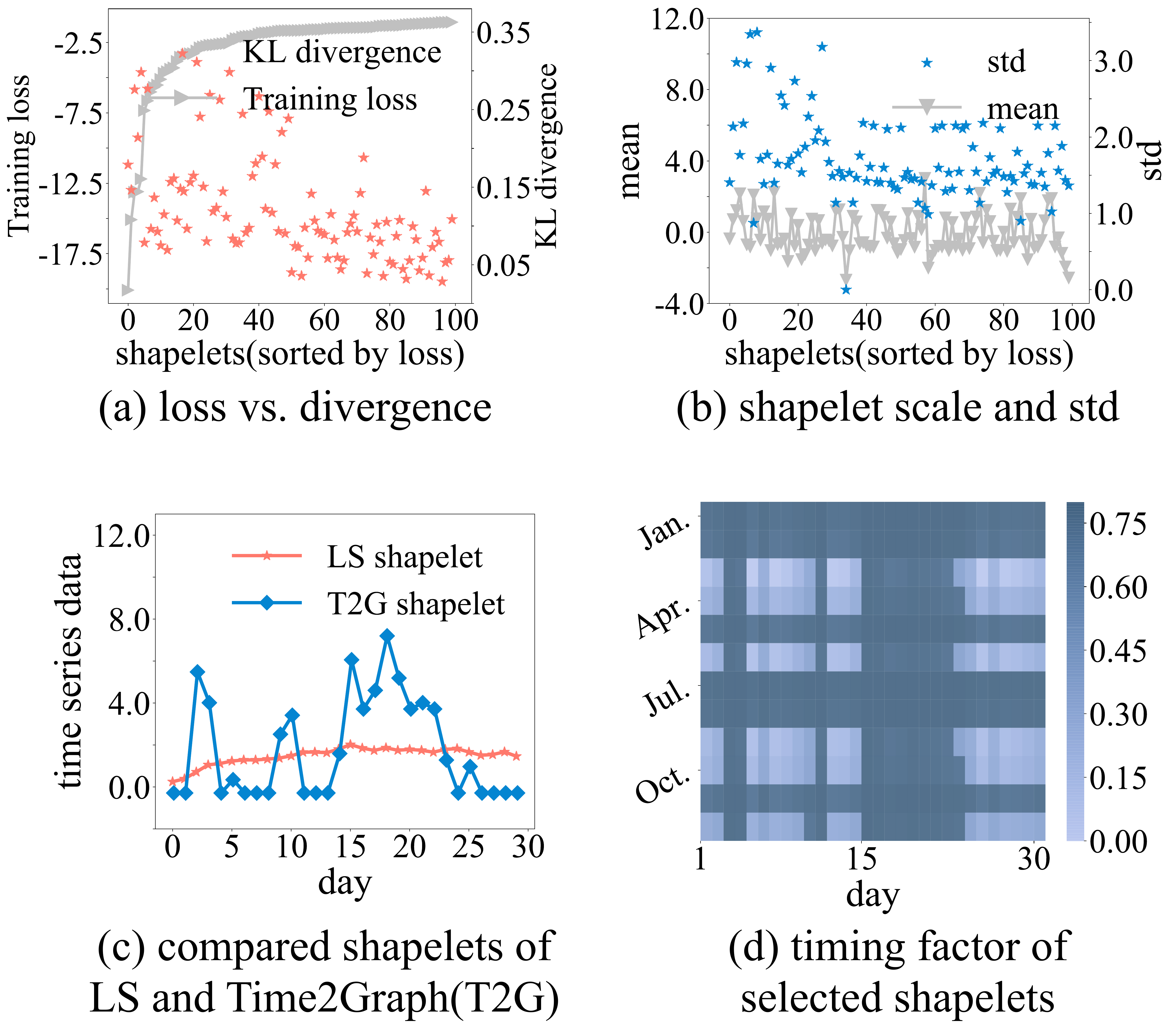}
		\end{subfigure}
		\caption{Shapelet analysis.
			\small 
			(a) shows the training and testing loss;
			(b) shows the mean and standard deviation (std) of shapelets;
			(c) compares the top (rank-1) shapelet between \textit{Learn Shapelets (LS)} and \textit{time2graph} (\textit{T2G}),
			and (d) visualizes the timing factors of the \textit{T2G}-shapelet in (c).
			\normalsize 
		}
		\label{fig:shapelet-detail}
		\vspace{-1em}
	\end{minipage}
\end{figure}

\subsection{Case Study of Time-Aware Shapelets}
\label{subsec:case:shapelet}
In the following two sections, we conduct several case studies to explore the interpretability of our proposed model,
and we use \sgdata dataset as the example since much domain knowledge is available here from experts.

The first question is, do the shapelets we extracted indeed have varying levels of discriminatory power?
As shown in \figref{fig:shapelet-detail}a, the training loss grows much slower at the right end, 
and the KL divergence of distributions of distances between positive and negative samples towards the top (ranked 1-50) shapelets on the test set is statistically significantly
 larger than that for the bottom (ranked 51-100) shapelets ($p=7.7*10^{-6}$). 
This reflects the effectiveness of the selected shapelets to some extent. 
To rigorously check the differences in shapelet variance, we show each shapelet's mean value and standard deviation (std) in \figref{fig:shapelet-detail}b.
Again, the std of top shapelets are statistically significantly larger than those of bottom ones ($p=7.5*10^{-3}$) ,
while the mean values across shapelets exhibit very little difference; 
this suggests that typical patterns intend to be unstable.
And to make further illustration, we compare the top-1 shapelet extracted by \textit{LS} (a popular baseline) and \textit{Time2Graph} in \figref{fig:shapelet-detail}c, d.
The scale and trends of these two shapelets differ a lot, and \figref{fig:shapelet-detail}d provides additional information towards time-aware shapelets in \textit{T2G}:
this specific shapelet matters in spring and summer (from month-level timing factor), and weights more at the peak of time series data during the month (from day-level  timing factor).
Such clue is the distinct advantage of our proposed model on the interpretability.

\subsection{Case Study of the Shapelet Evolution Graph}
We finally conduct experiments to construct shapelet evolution graphs for different time steps in order to see how the graphs change and how the nodes evolve. 
\figref{fig:graph:multi-graph} shows two graphs, one for January and another for July. 
In January, shapelet \textit{\#45} has large in/out degrees, 
and its corresponding timing factor is highlighted in January and February (dark areas). 
It indicates that shapelet \textit{\#45} is likely to be a common pattern at the beginning of a year.
As for July, shapelet \textit{\#45} is no longer as important as it was in January. 
Meanwhile, shapelet \textit{\#42}, which is almost an isolated point in January, becomes very important in July. 
Although we do not explicitly take seasonal information into consideration when constructing shapelet evolution graphs,
the inclusion of the timing factors means that they are already incorporated into the process of the graph generation.

%% file: relate.tex
\section{Related Work}
\label{sec:relate}
Time series modeling have attracted extensive research over a wide range of fields, such as 
image alignment~\cite{peng2014head}, 
speech recognition~\cite{shimodaira2002dynamic}, 
and motion detection~\cite{seto2015multivariate}.
One important technique here is Dynamic Time Warping (DTW)~\cite{muller2007dynamic}, 
which aims to find the proper distance measurement between time series data,
and a wide range of applications\cite{jeong2011weighted,gorecki2013using} have been proposed based on this metric. 
Traditional time series classification models try to extract efficient features from original data and develop a well-trained classifier, such as BoP~\cite{lin2012rotation}, TSF~\cite{deng2013time},
EE~\cite{lines2015time}, etc.

The major challenge is that there are no explicit features in sequences~\cite{xing2010brief},
so much research has focused on time series embedding\cite{Bagnall2017}:
Algorithms based on DTW and traditional embedding techniques~\cite{hayashi2005embedding} 
aim to project original time series data into feature-vector space;
Symbolic representations~\cite{lin2007experiencing,lin2003symbolic,schafer2015boss}  
transform time series using symbols such as characters in a given alphabet;
Shapelet discovery-based models~\cite{ye2011time,lines2012shapelet,rakthanmanon2013fast,lu16efficient,baydogan2016time},
from another perspective, try to find typical subsequences based on certain criteria such as information gain. 
Another relevant work to this paper 
is graph embedding, and one popular category\cite{goyal2018graph} lies in the random walk-based methods, 
such as DeepWalk~\cite{perozzi2014deepwalk} and node2vec~\cite{grover2016node2vec}.

%% file: conclude.tex
\section{Conclusion and Discussion}
\label{sec:conclude}
In this paper, we proposed a novel framework \modelname~to learn time-aware shapelets
for time series representations.
Moreover, to capture the co-occurrence and peer influence between shapelets, 
we put forward the idea of considering a time series as a graph, 
in which the nodes refer to shapelets,  
and weighted edges denote transitions between shapelets with varying probabilities.
By conducting experiments on three public datasets from \textit{UCR Archive} and two real-world datasets, 
we demonstrate the effectiveness and interpretability of our proposed model. 

%% file: appendix.tex
\clearpage
\appendix 
\section{Appendix}
\label{appendix}

\subsection{Algorithm Details in Our Approach}
\label{subsec:appendix:alg}
Here we illustrate the details of some proposed  algorithms.

\vpara{Shapelet candidate generation algorithm.}
We learn time-aware shapelets from a pool of candidates.
To reduce the size of the shapelet candidate pool for algorithm efficiency,
we apply a greedy strategy to select shapelet candidates from all possible subsequences.
The key idea is to maximize the peer distances between selected shapelets, thereby to fulfill the candidate searching space.
See \algref{alg:candidate} for details.
\begin{algorithm}[h]
	\caption{Shapelet Candidate Generation}
	\label{alg:candidate}		
	\begin{algorithmic}[1]
		\REQUIRE ~~\\
		time series data $T$, length $l$ and candidate size $|\mathcal{C}|$\\
		\ENSURE ~~\\
		Set of shapelets candiates $\mathcal{C}$\\
		\label{alg2:output}
		\STATE $seq \leftarrow$ all subsequences with length $l$ in $T$
		\label{alg2:init}
		\STATE Initialize $dist$ as a zero vector of size len($seq$),
		\\with a $1$ in the index of the closest subsequence\\ towards the center of $seq$
		\FOR{$cnt = 1, 2\cdots K$}
		\STATE $i \leftarrow$ argmax($dist$)
		\STATE $\mathcal{C}$.add($seq[i]$) and set $dist[i]$ as $-1$
		\STATE $update\_idx$ = argwhere($dist != -1$)
		\FOR{$j$ in $update\_idx$}
		\STATE $dist[j] \pluseq d(seq[i], seq[j])$ by Euclidean distance
		\ENDFOR
		\ENDFOR
		\RETURN $\mathcal{C}$
	\end{algorithmic}
\end{algorithm}

\vpara{Time-aware shapelet extraction algorithm.}
Based on the selected shapelet candidates generated by \algref{alg:candidate},
we then extract time-aware shapelets, 
where details and formulations of the extraction process are referred in \algref{alg:timing factor}.
\begin{algorithm}[h]
	\caption{Time-Aware Shapelet Extraction}
	\label{alg:timing factor}		
	\begin{algorithmic}[1]
		\REQUIRE ~~\\
		time series data $T=\{\bm{t}_1, \bm{t}_2\cdots\bm{t}_n\}$,\\
		number of shapelets to be discovered $K$ with length $l$,\\
		hyperparameters $\lambda, \epsilon$, and candidate size $|\mathcal{C}|$
		\ENSURE ~~\\
		$K$ time-aware shapelets with corresponding \\timing factor $\bm{w}_i, \bm{u}_i, \mbox{for } 1\le i \le K$\\
		\label{alg1:output}
		\STATE $\mathcal{C} \leftarrow$ GenerateShapeletCandidate($T, l, |\mathcal{C}|$) in \algref{alg:candidate}
		\label{alg1:init}		
		\FOR{$\bm{v}_i$ in $\mathcal{C}$}
		\STATE Initialize $\bm{w}_i$ and $\bm{u}_i$
		\STATE Optimize loss $\mathcal{\hat{L}}_i$ as \eqnref{eq:timing-loss} using \textit{Adam} optimizer
		\ENDFOR
		\RETURN $K$ shapelets and corresponding $\bm{w}_i, \bm{u}_i$ \\
		with minimum loss $\mathcal{\hat{L}}_i, 1\le i \le K$
	\end{algorithmic}
\end{algorithm}

\vpara{Greedy-DTW algorithm.}

\begin{figure}[h]
	\begin{minipage}{0.47\textwidth}
		\centering
		\includegraphics[width=\textwidth]{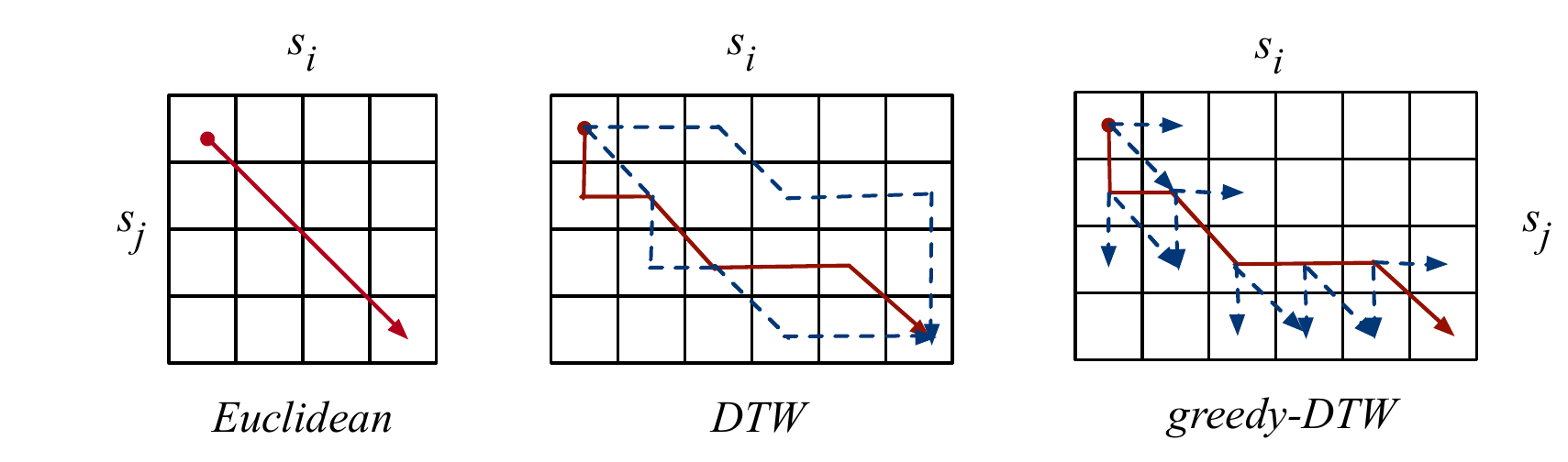}
		\caption{Illustration of Greedy-DTW distance}
		\label{fig:greedy-dtw}
	\end{minipage}
\end{figure}

One concern of \algref{alg:timing factor} is run-time efficiency, especially when we use \textit{DTW} as the distance metric.
So we adopt a algorithm named greedy-DTW to speed up the computation.
From the perspective of entry alignment, as shown in \figref{fig:greedy-dtw}, 
Euclidean distance (ED) is just the diagonal alignment for sequences with equal length $l$ ($\mathcal{O}(l)$),
while DTW aims to find a closest path from one corner to another.

The solution to the optimal path can be implemented by Dynamic Programming ($\mathcal{O}(l_1\cdot l_2)$),
and it is straightforward to develop a greedy algorithm to reduce the time complexity.
More precisely, instead of searching for global optimum,  
we can determine the alignment path at each step to achieve a local optimum, 
which is illustrated in \figref{fig:greedy-dtw} as \textit{greedy-DTW}. 
It is almost a linear search, 
as its time complexity decreases to $\mathcal{O}(l_1 + l_2)$ where the lengths of compared sequences are $l_1$ and $l_2$ respectively.
To avoid extremely asymmetrical alignment, we can set a sliding window to control the shifts.
See formulations of the greedy-DTW in \algref{alg:greedy-DTW}.

\begin{algorithm}[h]
	\caption{Greedy-DTW}
	\label{alg:greedy-DTW}		
	\begin{algorithmic}[1]
		\REQUIRE ~~\\
		compared sequence $\bm{s}_1,\bm{s}_2$ with length $l_1, l_2$,\\
		element-wise distance metric $\tau(\cdot,\cdot)$, \\
		sliding window $\omega$ for the timing shift 
		\ENSURE ~~\\
		greedy-DTW distance $\hat{d}$\\
		\STATE $\hat{d}, idx\_1, idx\_2 \leftarrow 0, 1, 1$ 
		\WHILE{$idx\_1 < l_1$ or $idx\_2 < l_2$}
		\STATE $dist\_1 \leftarrow \tau(\bm{s}_1(idx\_1 + 1), \bm{s}_2(idx\_2))$ 
		\STATE $dist\_2 \leftarrow \tau(\bm{s}_1(idx\_1), \bm{s}_2(idx\_2 + 1))$ 
		\STATE $dist\_3 \leftarrow \tau(\bm{s}_1(idx\_1 + 1), \bm{s}_2(idx\_2 + 1))$
		\STATE Warp for the direction where $dist\_i$ is the minimum
		\IF{$|idx\_1 - idx\_2| \ge \omega$}
		\STATE Roll back the last step and warp to the opposite direction
		\ENDIF
		\STATE $\hat{d} \leftarrow \hat{d} + \tau(\bm{s}_1(idx\_1), \bm{s}_2(idx\_2))$
		\ENDWHILE
		\STATE Warp for remaining elements along the unfilled direction
		
		\RETURN $\hat{d}$
	\end{algorithmic}
\end{algorithm}

\subsection{Implementation Details}
\label{subsec:implement}
Since intrinsic properties related to the scale and variance of the original time series may be lost in the process of shapelet extraction and embedding, 
we always incorporate handcraft features (including the mean and std of time series data within segments) with embedding vectors; this is what the \modelname~model exactly refers to.

One potential issue with \modelname concerns its efficiency.
Defining the number of time series as $|T|$, and each time series can be divided into $m$ segments with equal length of $l$;
then the time complexity of serially computing the loss defined in \eqnref{eq:timing-loss} for $|\mathcal{C}|$ shapelet candidates would be $\mathcal{O}(|T||\mathcal{C}|ml^2)$.
This figure is untenable when $l$ increases, or when $|T|$ and $|\mathcal{C}|$ are too large,
and both of them may be common in real-world applications.
Note that the square term $l^2$ is caused by DTW computation which is computationally costly, 
and it is possible to optimize this term from the algorithm level.
In practice, we use greedy-DTW (see in \algref{alg:greedy-DTW}) to replace original DTW metric, 
reducing the time complexity to $\mathcal{O}(l_1 + l_2)$ where the lengths of compared sequences are $l_1$ and $l_2$ respectively.

As to the implementation of the \modelname~model, we choose greedy-DTW as the distance metric in \eqnref{eq:w-seg-distance} as time warping (shifts) is an important characteristic of time series data, 
use \textit{softmin} to replace $\min$ in \eqnref{eq:s-t timing distance} to make the object function in \eqnref{eq:timing-loss} differentiable,
and select KL divergence as the function $g$ in \eqnref{eq:timing-loss}.
Specifically, we assume that peer distances between a shapelet candidate and both positive and negative samples follow a Gaussian distribution,
denoted by $\mathcal{N}_{pos}(\mu_{pos}, \sigma^2_{pos})$ and $\mathcal{N}_{neg}(\mu_{neg}, \sigma^2_{neg})$ respectively,
where the distribution parameters and their gradients can be easily computed by basic statistics.

There are two existing algorithms used in \modelname~: the graph embedding model (\textit{DeepWalk}) and outer classifier (\textit{XGBoost}).
Of course, there are multiple choices for these two parts, and which one to use is not the key component of the proposed model.  
But indeed we have several reasons to make our final decision.
For the graph embedding algorithm, \textit{DeepWalk} is a simple but strong baseline in the literature, 
and such random-walk-based model is expected to capture the evolution path of nodes (shapelets).
And for the choice of outer classifier,
we choose \textit{XGBoost} 
since it is often the winning solution in Kaggle competitions and shows strong performance compared to other methods such as Logistic Regression.
Besides, the ablation experiment in \tableref{tb:exp:performance} shows that 
the proposed representation outperforms handcraft features under the same classifier,
excluding the assumption that the performance incrementation is mainly brought by the outer classifier.

Finally, in the evaluation experiments, we use 5-fold nested cross-validation to conduct fine-tuning on hyperparameters 
(the public datasets are already split into training and testing sets, so we split validations sets on training samples), 
and compare classification performance on test sets for all public and the two real-world datasets.
Other settings for hyperparameters of the \modelname~model, as well as \textit{DeepWalk} and \textit{xgboost}, can be found in  \secref{subsec:appendix-para}.

\subsection{Hyperparameter Settings}
\label{subsec:appendix-para}

We have discussed several important hyperparameter settings of baseline methods and the proposed model in \secref{subsec:exp setup} and \ref{subsec:implement}. 
The remaining parts of parameter options are introduced below for better reproducibility.

\vpara{Hyperparameters in baselines.}

Note that we use the source codes for most of the baseline methods provided by Bagnall, et.al\cite{Bagnall2017} on the website \url{http://www.timeseriesclassification.com},
except for LPS (use R codes provided by the original authors\cite{baydogan2016time})
and deep models (self-coded).
For distance-based and feature-based models, we use the default parameters, and for shapelet-based models, 
we set the number of shapelets and segment length as optimal values we find in the \modelname~model if the parameter interface is open.

\vpara{Hyperparameters in \modelname.}
\begin{itemize}[leftmargin=*]
	\item \textit{DeepWalk}: We only tune the representation-size, or saying, graph embedding size,
	and others are set as default;
	\item \textit{XGBoost}: We conduct grid search on the hyperparameters of \textit{xgboost} as follows:
	the maximum depth ({1, 3, 5, 7, 9}), learning rate ({0.1, 0.2}), 
	and class weight ({1, 10, 50, 100}). Others parameters are set as default.
	\item \modelname:
	We set the hyperparameters of the proposed model also by grid search at the number of shapelets $K$ and segment length $l$ 
	(number of segments $m$ and the segment length $l$ are bindingly determined since the total length of time series is fixed), while the search space may differ between different datasets. The number of shapelet candidates $|\mathcal{C}|$ is set as 100 times of $K$, the percentile for distance threshold $\delta$ is set as 10,
	and penalty parameters $\lambda$ and $\epsilon$ in \eqnref{eq:timing-loss} are fixed as 0.5 and 0.1 respectively.
	In batch-wise training for time-aware shapelets,  the batch size is set as 50, and we choose \textit{Adam} as loss optimizer.
	
\tableref{tb:appendix:paras} lists the hyperparameters of the \modelname~model for all datasets in our experiments.
\end{itemize}
\begin{table}[h]
	\centering
	\renewcommand\arraystretch{1.1}
	\addtolength{\tabcolsep}{-1.2pt}
	\begin{tabular}{l|ccccc}
		\toprule
		\multirow{2}{*}{\textbf{\diagbox[width=3cm]{Parameter}{Dataset}}} & \multirow{2}{*}{\textbf{\ucreqk}} & \multirow{2}{*}{\textbf{\ucrworms}}  &\multirow{2}{*}{\textbf{\ucrstra}} & \multirow{2}{*}{\textbf{\sgdata}} & \multirow{2}{*}{\textbf{\teledata}} \\
		
		& & & & &\\
		\midrule
		latent dimension & \multirow{2}{*}{32} &  \multirow{2}{*}{128} &  \multirow{2}{*}{256}&  \multirow{2}{*}{256} &  \multirow{2}{*}{64}\\
		 in DeepWalk&  &  &  &  & \\
		\midrule
		maximum depth &8 & 12 & 4 & 4 & 8\\
		learning rate & 0.1 & 0.2 & 0.2 &  0.1 & 0.1\\
		class weight & 10 & 1 & 10 & 10 &10\\
		
		\midrule
		number of shapelets & 50 & 20 & 50 & 100 & 80\\
		number of segments & 21& 30 & 15 & 12 & 40\\
		segment length & 24 & 30 & 15 & 30 & 24\\
		\bottomrule
	\end{tabular}
	\normalsize
	\caption{
		Hyperparameter settings for the \modelname~model.
	}
	\label{tb:appendix:paras}
\end{table}

\vpara{Concerns related with hyperparameters tuning.}

One concern may raise as is the extra predictive power brought by parameter tuning rather than the model itself, 
since we use default parameter settings in some baselines.
It can be addressed from several perspectives.
First, as illustrated in \secref{subsec:parameter}, some hyperparameters in \modelname model, such as segment length and number of shapelets, 
should be determined and tuned carefully since they are highly co-related with the intrinsic characteristics of the dataset.
As for the outer classifier, we conduct nested cross-validation to select proper parameters, 
aiming at better performance as well as avoiding overfitting.
When it comes to baseline methods, there are two cases: 
for hyperparameters related with the dataset properties,  we use the same setting as in the \modelname to maintain consistency, 
i.e., number of shapelets and segment length for shapelet-based models;
for other model-specific parameters, if we can easily touch the interface, 
i.e., \textit{XGBoost}, we conduct fine-tuning to find best options, as we have done in \textit{XGBoost-origin} and \textit{-feature}.
Since distance-based models have very few parameters needed to adjust, 
and in most of our selected baselines, parameters have already been tuned on benchmarks,
it seems reasonable to use the default settings.

\subsection{Case Study of Typical Shapelets}
\begin{figure}[ht!]
	\begin{minipage}{0.47\textwidth}
		\centering
		\begin{subfigure}[t]{0.31\textwidth}
			\includegraphics[width=\textwidth]{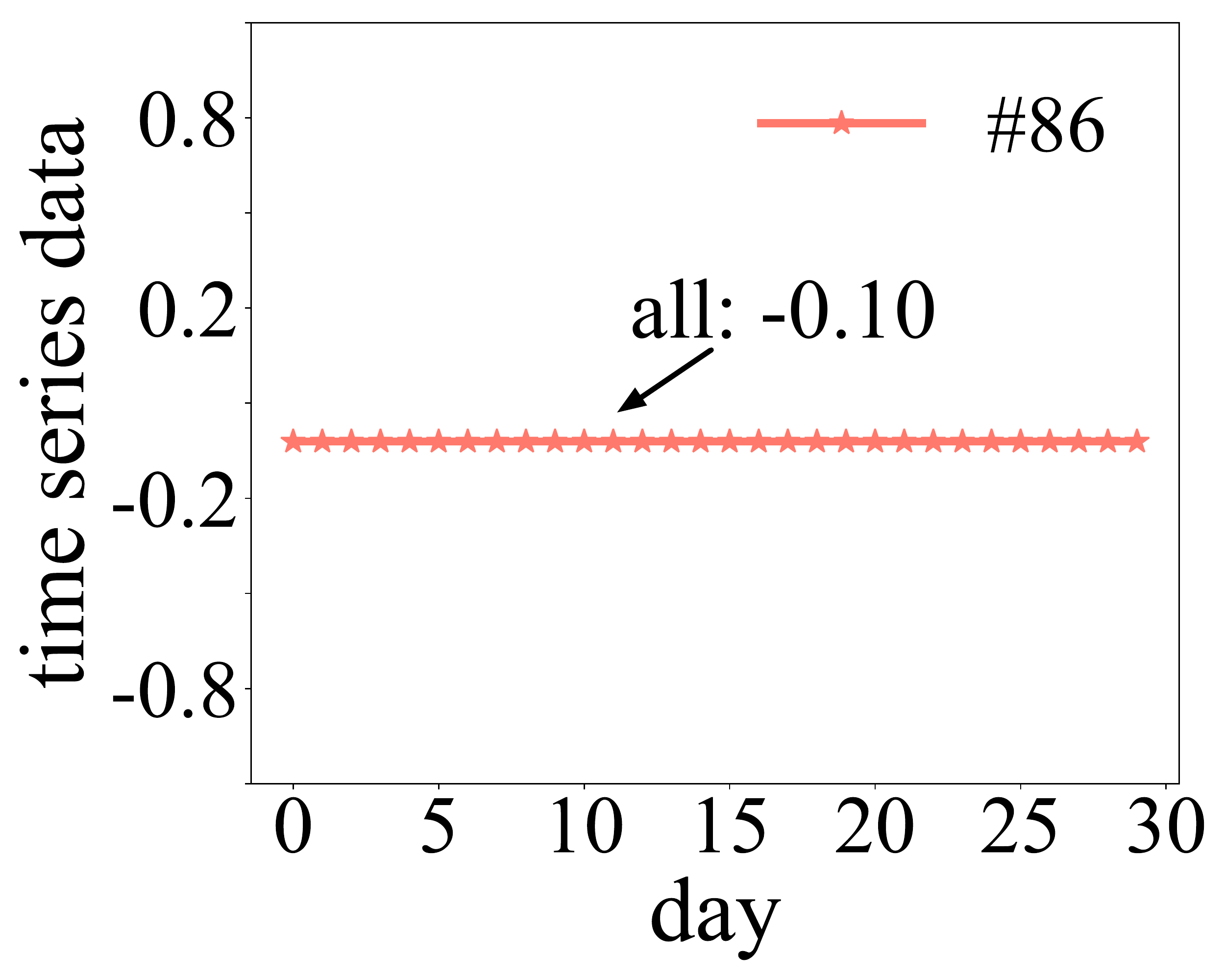}
			\caption{shapelets \#86}
			\label{fig:shapelet:86}
		\end{subfigure}
		\hfill
		\begin{subfigure}[t]{0.31\textwidth}
			\includegraphics[width=\textwidth]{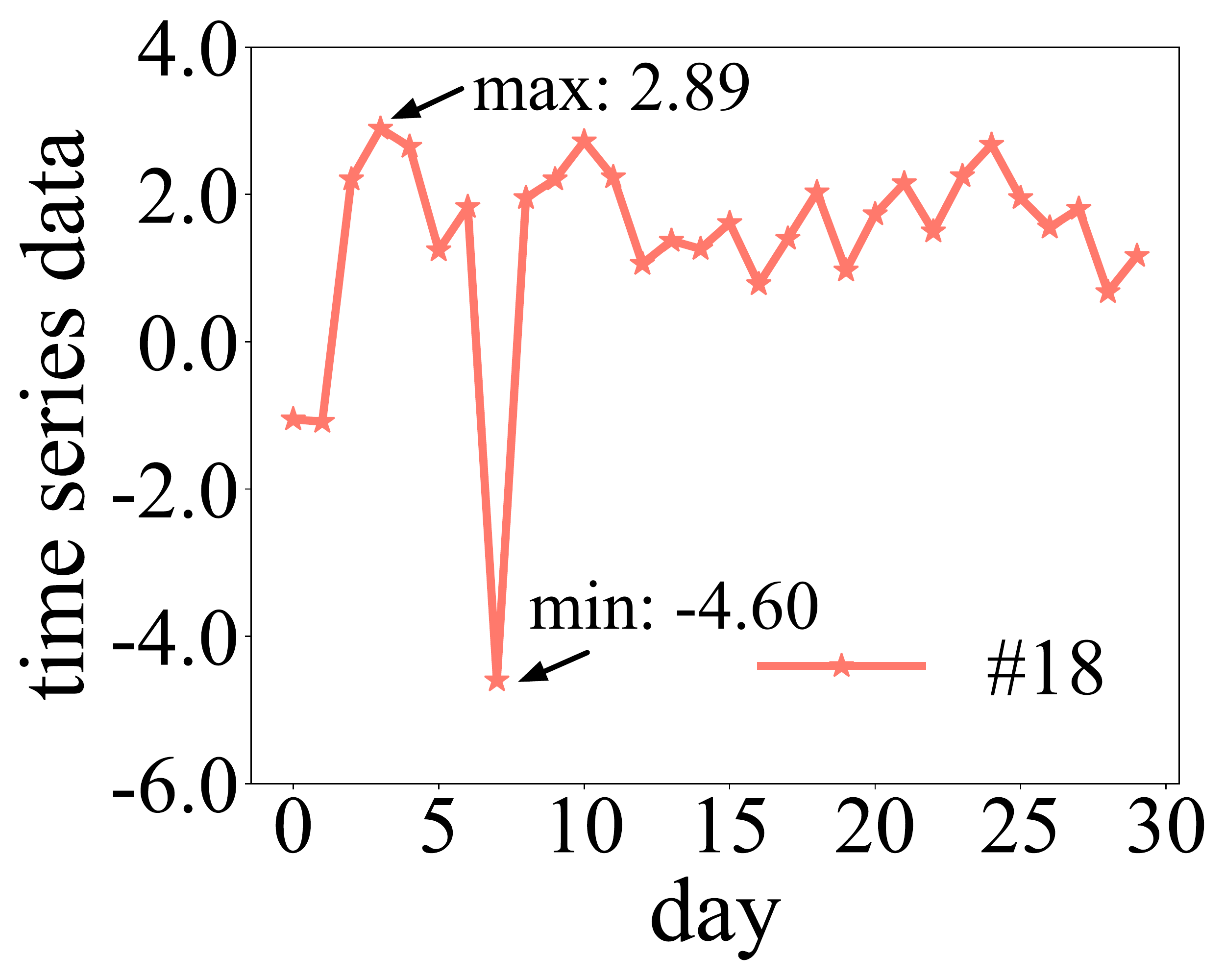}
			\caption{shapelets \#18}
			\label{fig:shapelet:18}
		\end{subfigure}
		\hfill
		\begin{subfigure}[t]{0.31\textwidth}
			\includegraphics[width=\textwidth]{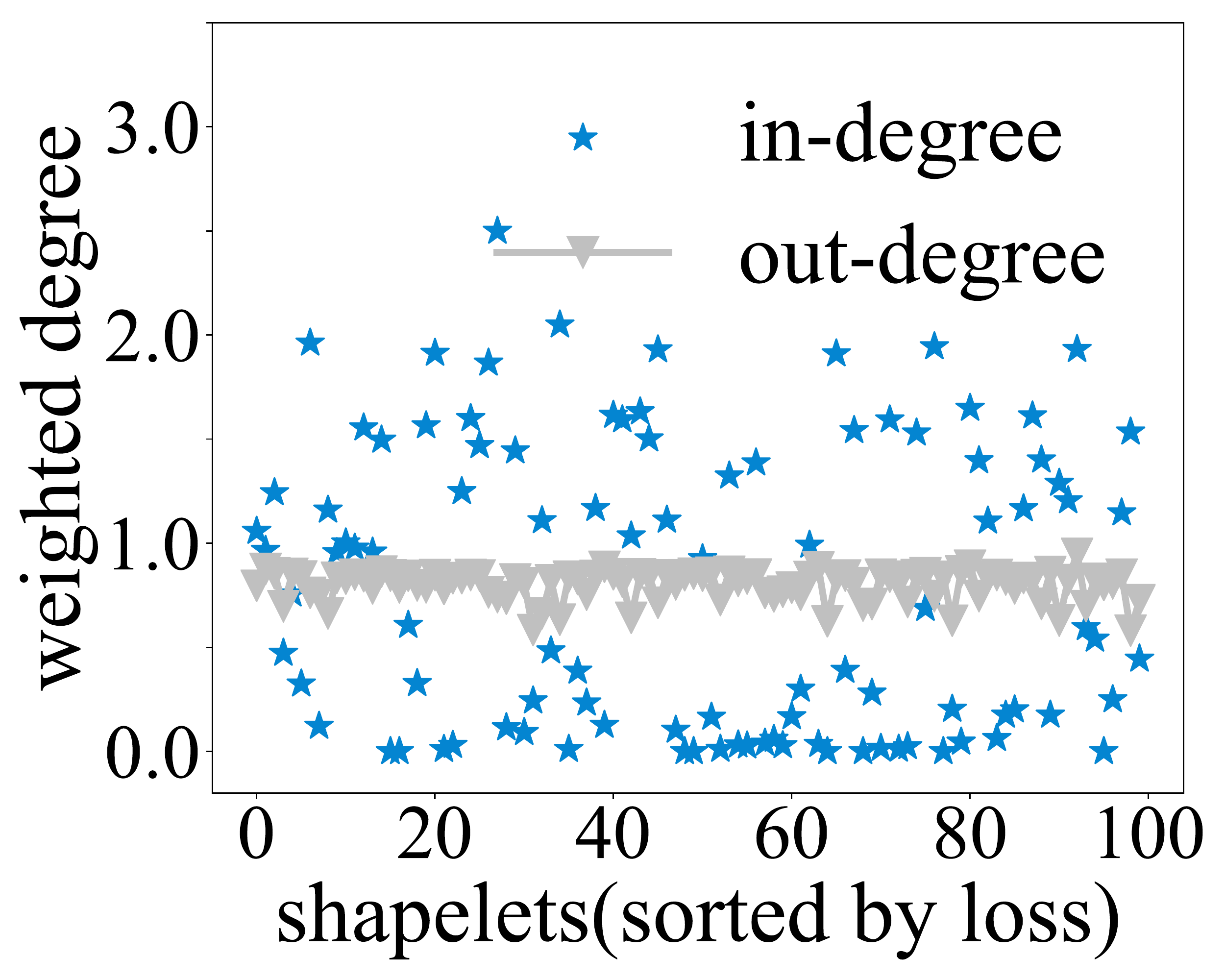}
			\caption{in/out degree}
			\label{fig:graph:loss-degree}
		\end{subfigure}
		\caption{Case study. 
			\small 
			(a) shows a shapelet with a stable scale.
			(b) shows a shapelet with an unstable scale.
			(c) illustrates that shapelet in-degree (i.e., the probability to be transferred into) varies a lot.
			\normalsize 
		}
		\label{fig:graph:uni-analysis}
		\vspace{-2pt}
	\end{minipage}
\end{figure}
\figref{fig:graph:unigraph} shows the shapelet evolution graph for the \sgdata dataset. 
It seems that shapelets such as \textit{\#78, \#43, and \#86} have higher degrees than most of the others,
while edges such as \textit{\#67} $\rightarrow$ \textit{\#85} are more important.
We show shapelet \textit{\#86} as an example in \figref{fig:shapelet:86}: 
Its scale is wholly unchanged ($0.10$), and is close to the median value (i.e., $0$, as we have proceeded with 0-1 normalization). 
Therefore, \#86 has more edges than others,
since it is close to other common vertices. 
As a comparison, \figref{fig:shapelet:18} shows shapelet \textit{\#18}, which has a low weighted degree.
We can see that its scale ranges from $-4.60$ to $2.89$, and its variance is relatively large.
Common sense dictates that it should seldom appear among normal users, so its degree would be rather small.

Another intuitive observation from \figref{fig:graph:unigraph} is that 
the node size in the graph varies significantly; in other words, the weighted in-degrees have a large range. 
We illustrate this finding in \figref{fig:graph:loss-degree}. 
The up-down fluctuations of in-degree indicate the random distribution ($p=0.07$),
while the out-degrees are stable.
This is reasonable, since each shapelet is almost bound to evolve into some other shapelets (including itself), 
but there exist nodes that are infrequently transferred into.
Such characteristics also explain that low in-degree nodes 
lead to completely isolated transition paths.

\subsection{Reclaim the Classification Result}
\label{subsec:appendix:reclaim}
It seems that our model performs better on real-world datasets compared with three public datasets, 
since the accuracy improvements on those \textit{UCR} datasets are marginal. 
It is caused by several reasons: 
1) Firstly, the sizes of datasets on \textit{UCR Archive} are small, 
which makes it  difficult to capture the shapelet evolution patterns 
(Generally we can assume that such transition patterns can be better extracted from large samples).
This may affect the performance of our proposed model.
2) Another characteristic of \textit{UCR Archive} datasets is that they are often well-balanced.
Since \modelname is designed to capture normal shapelet evolutions \textit{only among normal samples},
imbalanced setting (very few abnormal cases) would bring little negative effects on mining evolution dynamics.
This may explain why our model can outperform the baselines especially in the imbalanced scenarios.

\subsection{Public Dataset Description}
\label{subsec:appendix:dataset}
\vpara{Earthquakes (\ucreqk).}
The data are taken from the Northern California Earthquake Data Center. 
Each data point is an averaged reading for 1 hour, ranging from 1967 to 2003. 
Then it turned into a classification problem of differentiating between a positive and negative major earthquake event defined by domain experts.
In total, 368 negative and 93 (25.3\%) positive cases were extracted from 86,066 hourly readings. 

\vpara{WormsTwoClass (\ucrworms).}
Caenorhabditis elegans (C. elegans) is a roundworm commonly used as a model organism in genetics study. 
The data of the movement of these worms were formatted for the time series classification task. 
Each case is a series with the length of 900, and there are 258 cases in total;
in this task, the aim is to classify worms of wild-type (109 (42.2\%) samples) or mutant-type.

\vpara{Strawberry (\ucrstra).}
Food spectrographs are used in chemometrics to classify food types.
This data was processed using Fourier transform infrared spectroscopy with attenuated total reflectance sampling. 
There are a total of 983 time series samples with length 235, and the classes are strawberry purees (authentic, 632 (64.3\%) samples) and non-strawberry purees (adulterated strawberries and other fruits).

\subsection{Reproducibility}
\label{subsec:appendix-rep}
The public datasets from \textit{UCR Time Series Archive} can be downloaded from \url{https://www.cs.ucr.edu/~eamonn/time_series_data_2018/}.
\textit{UCR Archive} is the most widely used benchmark dataset for time series modeling. 
For evaluation and implementation convenience, we only consider binary classification (although the model is not limited to that), 
and finally choose three typical cases that used in our experiment.

Source codes including data preprocessing and model implementations can be found in \url{https://github.com/petecheng/Time2Graph}.
The \textit{DeepWalk} model is referred to \url{https://github.com/phanein/deepwalk}, 
with several minor modifications in the initialization of \textit{networkx} graph object  for weighed version.
The environment requirements and run instructions can be found in the \textit{README}  document.